\documentclass[10pt,twocolumn,letterpaper]{article}

\usepackage{cvpr}
\usepackage{times}
\usepackage{epsfig}
\usepackage{graphicx}
\usepackage{amsmath}
\usepackage{amssymb}

\usepackage[pagebackref=true,breaklinks=true,letterpaper=true,colorlinks,bookmarks=false,citecolor=blue,linkcolor=blue]{hyperref}

\cvprfinalcopy % *** Uncomment this line for the final submission

\def\cvprPaperID{1434} % *** Enter the CVPR Paper ID here

\ifcvprfinal\fi
\begin{document}

%%%%%%%%% TITLE
\title{Deep Image Harmonization}

\author{Yi-Hsuan Tsai$^{1}$
	\hspace{0.15in} Xiaohui Shen$^{2}$ \hspace{0.15in} Zhe Lin$^{2}$
	\hspace{0.15in} Kalyan Sunkavalli$^{2}$ \hspace{0.15in} Xin Lu$^{2}$
	\hspace{0.15in} Ming-Hsuan Yang$^{1}$
	\vspace{1mm} \\
	\hspace{0.1in} $^{1}$University of California, Merced \hspace{0.4in} $^{2}$Adobe Research
	\vspace{1mm} \\
	\hspace{0.1in} $^{1}${\tt\small \{ytsai2, mhyang\}@ucmerced.edu}
	\hspace{0.4in} $^{2}${\tt\small \{xshen, zlin, sunkaval, xinl\}@adobe.com}
}

\maketitle
%\thispagestyle{empty}

%%%%%%%%% ABSTRACT
\begin{abstract}
Compositing is one of the most common operations in photo editing.
To generate realistic composites, the appearances of foreground and background need to be adjusted to make them compatible.
Previous approaches to harmonize composites have focused on learning statistical relationships between hand-crafted appearance features of the foreground and background, which is unreliable especially when the contents in the two layers are vastly different.
In this work, we propose an end-to-end deep convolutional neural network for image harmonization, which can capture both the context
and semantic information of the composite images during harmonization.
We also introduce an efficient way to collect large-scale and high-quality training data that can facilitate the training process. 
Experiments on the synthesized dataset and real composite images show that the proposed network outperforms previous state-of-
the-art methods.
\end{abstract}

%%%%%%%%% BODY TEXT
\section{Introduction}
\begin{figure}[t]
	\centering
	\begin{tabular}
		{@{\hspace{0mm}}c@{\hspace{1mm}} @{\hspace{1mm}}c@{\hspace{0mm}}
		}
		\includegraphics[width=0.48\linewidth]{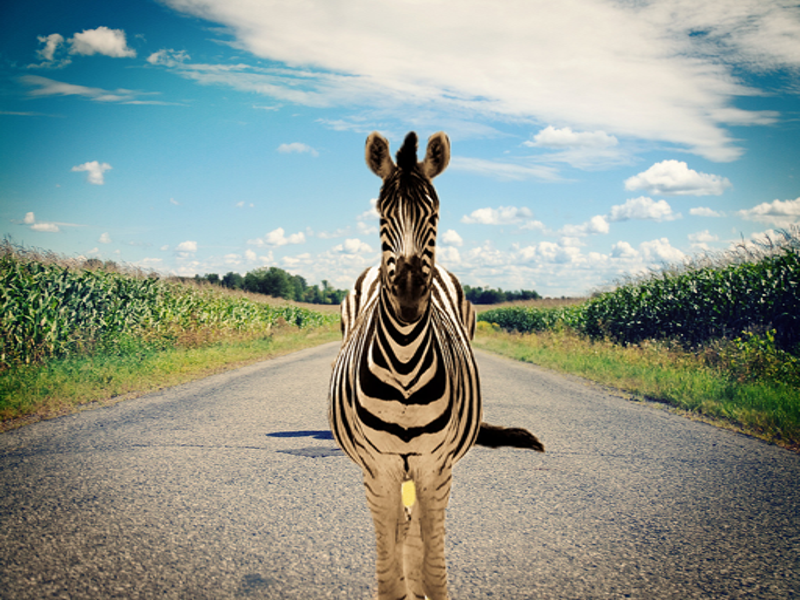} &
		\includegraphics[width=0.48\linewidth]{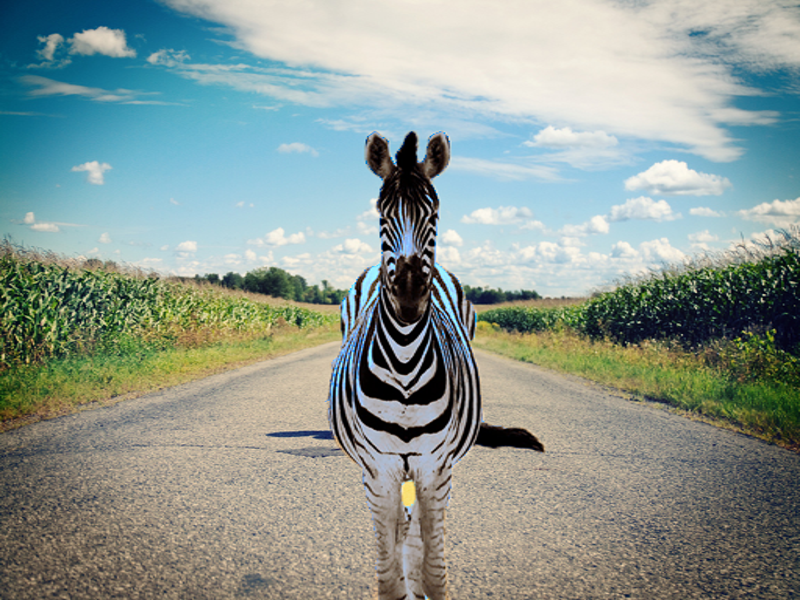} \\
		\vspace{1mm}
		Composite image & Xue \cite{Xue_siggraph_2012} \\
		\includegraphics[width=0.48\linewidth]{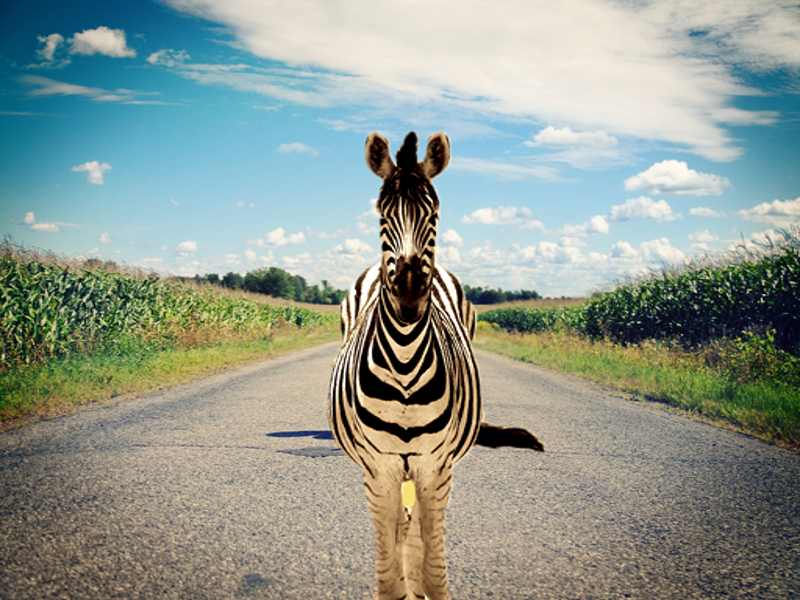} &
		\includegraphics[width=0.48\linewidth]{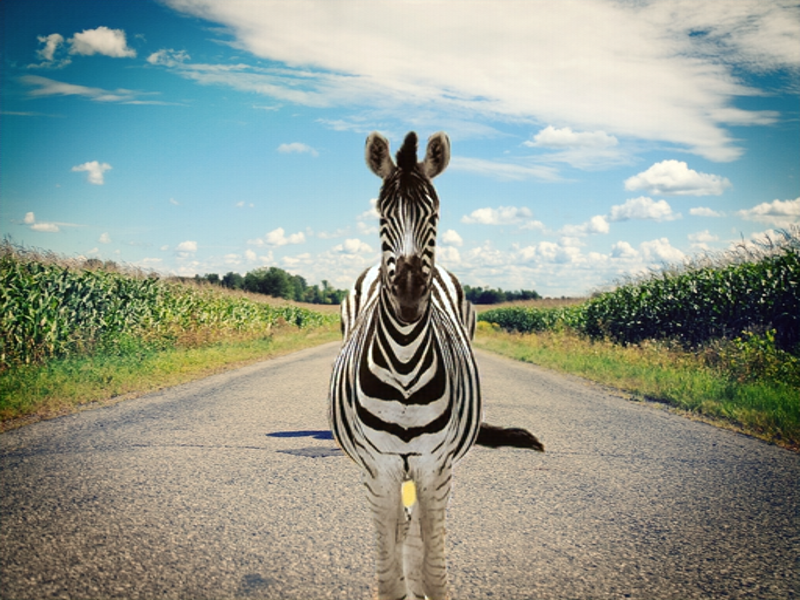} \\
		\vspace{1mm}
		Zhu \cite{Zhu_ICCV_2015} & Our harmonization result
	\end{tabular}
	\caption{Our method can adjust the appearances of the composite foreground to make it compatible with the background region.
	Given a composite image, we show the harmonized images generated by \cite{Xue_siggraph_2012}, \cite{Zhu_ICCV_2015} and our deep harmonization network.
	}
	\label{fig:intro}
%	\vspace{-5mm}
\end{figure}

Compositing is one of the most common operations in image editing. To generate a composite image, a foreground region in one image is extracted and combined with the background of another image.
However, the appearances of the extracted foreground region may not be consistent with the new background, making the composite image unrealistic.
Therefore, it is essential to adjust the appearances of the foreground region to make it compatible with the new background (Figure \ref{fig:intro}).
Previous techniques improve the realism of composite images by transferring statistics of hand-crafted features, including color \cite{Lalonde_ICCV_2007, Xue_siggraph_2012} and texture \cite{Sunkavalli_siggraph_2010}, between the foreground and background regions.
However, these techniques do not take the contents of the composite images into account, leading to unreliable results when appearances of the foreground and background regions are vastly different.

%
%Numerous methods have been proposed to improve the realism of composite images by using color compatibility \cite{Lalonde_ICCV_2007}, transferring appearance statistics \cite{Xue_siggraph_2012} and incorporating the realism model \cite{Zhu_ICCV_2015}.
%%
%For such harmonization methods, image statistics such as color, brightness and contrast are considered to match appearances between foreground and background regions.
%%
%However, such adjustment relies on the composite image itself or external exemplars, which may be unreliable in some cases, especially when appearances of the foreground and background regions are vastly different, or when there are no adequate exemplars.

%
In this work, we propose a learning-based method by training an end-to-end deep convolutional neural network (CNN) for image harmonization, which can capture both the context and semantic information of the composite images during harmonization.
Given a composite image and a foreground mask as the input, our model directly outputs a harmonized image, where the contents are the same as the input but with adjusted appearances on the foreground region.
Context information has been utilized in several image editing tasks, such as image enhancement \cite{Hwang_ECCV_2012, Yan_siggraph_2015}, image editing \cite{Tsai_SIGGRAPH_2016} and image inpainting \cite{Pathak_CVPR_2016}.
For image harmonization, it is critical to understand what it looks like in the surrounding background region near the foreground region.
Hence foreground appearances can be adjusted accordingly to generate a realistic composite image.
Toward this end, we train a deep CNN model that consists of an encoder to capture the context of the input image and a decoder to reconstruct the harmonized image using the learned representations from the encoder.
%
%For instance, when a bright foreground object is inserted into an image with a dark background, our encoder should learn the context from this background region and reconstruct the output image with the darker foreground object. 
%

In addition, semantic information is of great importance  to improve image harmonization.
%
%MH: check this to see whether I get it right or not
For instance, if we know the foreground region to be harmonized is a sky, 
it is natural to adjust the appearance and color to be blended with the surrounding contents, 
instead of making the sky green or yellow. 
However, the above-mentioned encoder-decoder does not explicitly model semantic information without the supervision of high-level semantic labels.
Hence, we incorporate another decoder to provide scene parsing of the input image, while sharing the same encoder for learning feature representations.
A joint training scheme is adopted to propagate the semantic information to the harmonization decoder.
With such semantic guidance, the harmonization process not only captures the image context but also understands semantic cues to better adjust the foreground region.

Training an end-to-end deep CNN requires a large-scale training set including various and high-quality samples.
%
%In addition to creating composite images, generating ground truth harmonized images requires professional editing skills and is not feasible to produce a large-scale set.
%
However, unlike other image editing tasks such as image colorization \cite{Zhang_ECCV_2016} and inpainting \cite{Pathak_CVPR_2016} where unlimited amount of training data can be easily generated, it is relatively difficult to collect a large-scale training set for image harmonization, as generating composite images and ground truth harmonized output requires professional editing skills and a considerable amount of time.
To solve this problem, we develop a training data generation method that can synthesize large-scale and high-quality training pairs, which facilitates the learning process.

To evaluate the proposed algorithm, we conduct extensive experiments on synthesized and real composite images.
We first quantitatively compare our method with different settings to other existing approaches for image harmonization on our synthesized dataset, where the ground truth images 
are provided.
We then perform a user study on real composite images and show that our model trained on the synthesized dataset performs favorably in real cases.

The contributions of this work are as follows.
First, to the best of our knowledge, this is the first attempt to have an end-to-end learning approach for image harmonization. 
Second, we demonstrate that our joint CNN model can effectively capture context and semantic information, and can be efficiently trained for both the harmonization and scene parsing tasks.
%First, we propose a learning-based method which captures both the context and semantic information for image harmonization.
%
%Second, we demonstrate that our joint CNN model can be efficiently trained in an end-to-end manner 
%
Third, an efficient method to collect large-scale and high-quality training images is developed to facilitate the learning process for image harmonization.
\section{Related Work}
%
%The goal of this work is to input a composite image and output a harmonized image where the content is the same as the input but with adjusted appearances on the foreground region.
Our goal is to harmonize a composite image by adjusting its foreground appearances while keeping the same background region.
In this section, we discuss existing methods closely related to this setting.
In addition, the proposed method adopts a learning-based framework and a joint training scheme. 
Hence recent image editing methods within this scope are also discussed.

{\flushleft {\bf Image Harmonization.}}
Generating realistic composite images requires a good match for both the appearances and contents between foreground and background regions.
Existing methods use color and tone matching techniques to ensure consistent appearances, such as transferring global statistics \cite{Reinhard_CGA_2001,Pitie_CVMP_2007}, applying gradient domain methods \cite{Perez_siggraph_2003,Tao_IJCV_2013}, matching multi-scale statistics \cite{Sunkavalli_siggraph_2010} or utilizing semantic information \cite{Tsai_SIGGRAPH_2016}.
While these methods directly match appearances to generate realistic composite images, realism of the image is not considered.
Lalonde and Efros \cite{Lalonde_ICCV_2007} predict the realism of photos by learning color statistics from natural images and use these statistics to adjust foreground appearances to improve the chromatic compatibility.
On the other hand, a data-driven method \cite{Johnson_TVGG_2011} is developed to improve the realism of computer-generated images by retrieving a set of real images with similar global layouts for transferring appearances.

In addition, realism of the image has been studied and used to improve the harmonization results.
Xue et al. \cite{Xue_siggraph_2012} perform human subject experiments to identify most significant statistical measures that determine the realism of composite images and adjust foreground appearances accordingly.
Recently, Zhu et al. \cite{Zhu_ICCV_2015} learn a CNN model to predict the realism of a composite image and incorporate the realism score into a color optimization function for appearance adjustment on the foreground region.
Different from the above-mentioned methods, our end-to-end CNN model directly learn from pairs of a composite image as the input and a ground truth image, which ensures the realism of the output results.  

{\flushleft {\bf Learning-based Image Editing.}}
Recently, neural network based methods for image editing tasks such as image colorization \cite{Iizuka_SIGGRAPH_2016, Larsson_ECCV_2016, Zhang_ECCV_2016}, inpainting \cite{Pathak_CVPR_2016} and filtering \cite{Liu_ECCV_2016}, have drawn much 
attention due to their efficiency and impressive results.
Similar to autoencoders \cite{Bengio_PAMI_2013}, these methods adopt an unsupervised learning scheme that learns feature representations of the input image, where raw data 
is used for supervision.
Although our method shares the similar concept, to the best of our knowledge
it is the first end-to-end trainable CNN architecture designed for image harmonization.

However, these image editing pipelines may suffer from missing semantic information in the finer level during reconstruction, and such semantics are important cues for understanding image contents.
Unlike previous methods that do not explicitly use semantics, we incorporate an additional model to predict pixel-wise scene parsing results and then propagate this information to the harmonization model, where the entire framework is still end-to-end trainable.
\section{Deep Image Harmonization}
\begin{figure*}[t]
	\centering
	\includegraphics[width=1.0\linewidth]{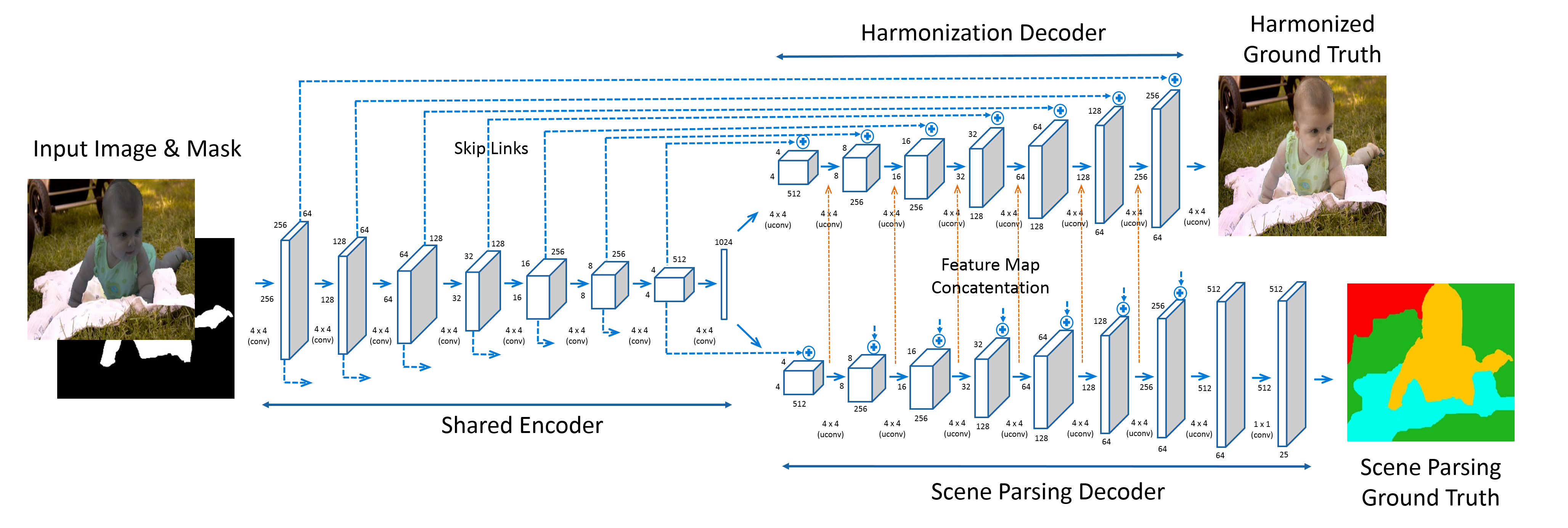}
	\caption{The overview of the proposed joint network architecture.
	Given a composite image and a provided foreground mask, we first pass the input through an encoder for learning feature representations.
	The encoder is then connected to two decoders, including a harmonization decoder for reconstructing the harmonized output and a scene parsing decoder to predict pixel-wise semantic labels.
	In order to use the learned semantics and improve harmonization results, we concatenate the feature maps from the scene parsing decoder to the harmonization decoder (denoted as dot-orange lines).
	In addition, we add skip links (denoted as blue-dot lines) between the encoder and decoders for retaining image details and textures.
	Note that, to keep the figure clean, we only depict the links for the harmonization decoder, while the scene parsing decoder has the same skip links connected to the encoder.
	}
	\label{fig:network}
\end{figure*}

In this section, we describe the details of our proposed end-to-end CNN model for image harmonization.
Given a composite image and a foreground mask as the input, our model outputs a harmonized image by adjusting foreground appearances while retaining the background region.
Furthermore, we design a joint training process with scene parsing to understand image semantics and thus improve harmonization results.
Figure \ref{fig:network} shows an overview of the proposed CNN architecture. 
Before describing this network, we first introduce a data collection method that allows us to obtain large-scale and high-quality training pairs.
%%
%\begin{figure}[t]
%	\centering
%	\includegraphics[width=1.0\linewidth]{figure/data_collection.png}
%	\caption{Overall process to generate input and output pairs for training the proposed network.
%		Given an original image as the ground truth, we first select a foreground region, 
%		and edit the appearances of this region to generate the synthesized composite image as the input to our network.	
%	}
%	\label{fig:data}
%%	\vspace{-2mm}
%\end{figure}
%
\subsection{Data Acquisition}
\label{sec:data}
Data acquisition is an essential step to successfully train a CNN.
As described above, an image pair containing the composite and harmonized images is required as the input and ground truth for the network.
Unlike other unsupervised learning tasks such as \cite{Zhang_ECCV_2016, Pathak_CVPR_2016} that can easily obtain training pairs, image harmonization task requires expertise to generate a high-quality harmonized image from a composite image, which is not feasible to collect large-scale training data.

\begin{figure}[t]
	\centering
	\includegraphics[width=1.0\linewidth]{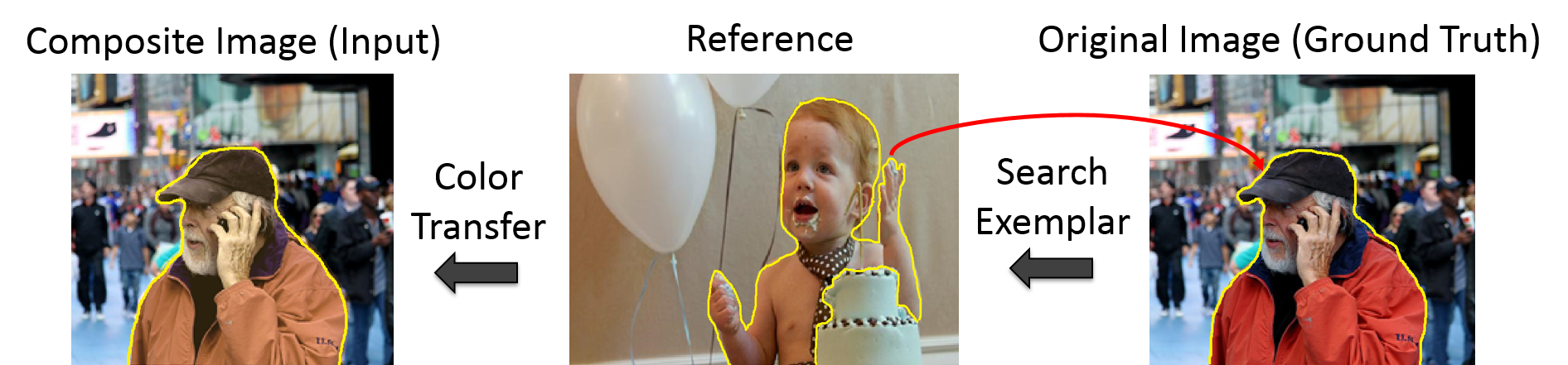}
	\vspace{2mm}
	(a) Miscrosoft COCO \& Flickr \\
	\includegraphics[width=1.0\linewidth]{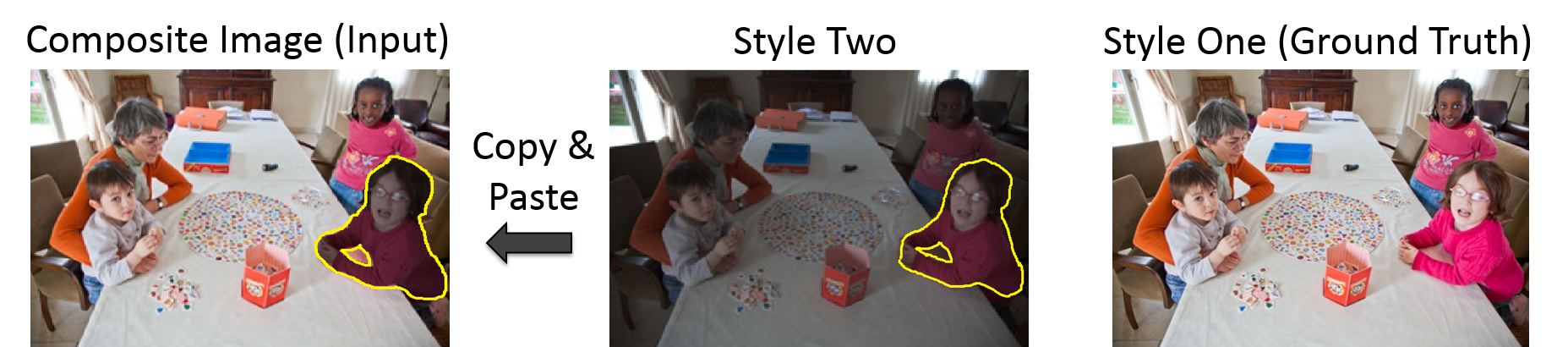}
	\vspace{1mm}
	(b) MIT-Adobe FiveK \\
	\caption{Data acquisition methods. We illustrate the approaches for collecting training pairs for the datasets (a) Miscrosoft COCO and Flickr via color transfer, and (b) MIT-Adobe FiveK with different styles.
	}
	\label{fig:data_coco_5k}
\end{figure}

To address this issue, we 
start from a real image which we treat as the output ground truth of our network.
We then select a region (e.g., an object or a scene) and edit its appearances to generate an edited image which we use as the input composite image to the network.
The overall process is described in Figure \ref{fig:data_coco_5k}.
This data acquisition method ensures that the ground truth images are always realistic 
so that the goal of the proposed CNN is to directly reconstruct a 
realistic output from a composite image.
In the following, we introduce the details of how we generate our synthesized dataset.
{\flushleft {\bf Images with Segmentation Masks.}}
We first use the Microsoft COCO dataset \cite{Lin_ECCV_2014}, 
where the object segmentation masks are provided for each image.
%
%MH: what color transfer method, cite it. ok,, you cite it at the end
To generate synthesized composite images, we randomly select an object and edit its appearances via a color transfer method.
%
%MH: check this sentence
In order to ensure that the edited images are neither arbitrary nor unrealistic in 
color and tone, 
we construct the color transfer functions by searching for proper reference objects.
%Here, to prevent that the edited image is completely corrupted due to the appearance changes in the object region, we carefully search proper reference objects for color transfer.
%

Specifically, given a target image and its corresponding object mask, we search a reference image which contains the object with the same semantics.
We then transfer the appearance from the reference object to the target object.
As such, we ensure that the edited object still looks plausible but does not match the background context.
For color transfer, we compute statistics of the luminance and color temperature, 
and use the histogram matching method  \cite{Lee_CVPR_2016}. 

To generate a larger variety of transferred results, we apply different transfer parameters for both the luminance and color temperature on one image, so that our learned network can adapt to different scenarios in real cases.
In addition,  we apply an aesthetics prediction model \cite{Kong_ECCV_2016} to filter out low-quality images.
An example of generated synthesized input and output pairs are shown in Figure \ref{fig:data_coco_5k}(a).
{\flushleft {\bf Images with Different Styles.}}
Although the Microsoft COCO dataset provides us with rich object categories, it is still limited to certain objects.
To cover more object categories, we augment it with the MIT-Adobe FiveK dataset \cite{Bychkovsky_CVPR_2011}.
In this dataset, each original image has another $5$ different styles that are re-touched by professional photographers using Adobe Lightroom, resulting in $6$ editions of the same image.
To edit the original image, we begin with one randomly selected style and manually segment a region.
We then crop this segmented region and overlay on the image with another style to generate the synthesized composite image.
An example set is presented in Figure \ref{fig:data_coco_5k}(b).
\begin{table}[t]
	\caption{Number of training and test images on three synthesized datasets.
	}
	\vspace{1mm}
	\small
	\centering
	\renewcommand{\arraystretch}{1.5}
	\setlength{\tabcolsep}{6pt}     
	\begin{tabular}{|c|c|c|c|}
		\hline
		& MSCOCO & MIT-Adobe & Flickr \\
		\hline
		
		Training set & 51187 & 4086 & 4720 \\
		\hline
		
		Test set & 3842 & 68 & 96 \\
		
		\hline
	\end{tabular}
	\label{tab:data}
	%	\vspace{-5mm}
\end{table}
{\flushleft {\bf Flickr Images with Diversity.}}
Since images in the MIT-Adobe FiveK and Microsoft COCO datasets only contain certain scenes and styles, we collect a dataset from Flickr with larger diversity such as images containing different scenes or stylized images. 
To generate input and ground truth pairs, we apply the same color transfer technique described for the Microsoft COCO dataset.
However, since there is no semantic information provided in this dataset to search proper reference objects for transfer, we use a pre-trained 
scene parsing model \cite{Zhou_corr_2016} to predict semantic pixel-wise labels.
We then compute a spatial-pyramid label histogram \cite{Lazebnik_CVPR_2006} of the target image and retrieve reference images from the ADE20K dataset \cite{Zhou_corr_2016} with similar histograms computed from the ground truth annotations.

Next, we manually segment a region (e.g., an object or a scene) in the target image.
Based on the predicted scene parsing labels within the segmented target region, we find a region in the reference image that shares the same labels as the target region.
The composite image is then generated by the color transfer method mentioned above (Figure \ref{fig:data_coco_5k}(a)).

{\flushleft {\bf Discussions.}}
With the above-mentioned data acquisition methods on three datasets, we are able to collect large-scale and high-quality training and test pairs (see Table \ref{tab:data} for a summarization).
This enables us to train an end-to-end CNN for image harmonization with several benefits.
First, our data collection method ensures that the ground truth images are realistic, so the network can really capture the image realism and adjust the input image according to the learned representations.

Another merit of our method is to enable quantitative evaluations.
This is, we can use the synthesized composite image to measure errors by comparing to the ground truth images.
Although there should be no single best solution for the image harmonization task, this quantitative measurement can give us a sense of how closer the images generated by different methods are, to a truly realistic image (discussed in Section \ref{sec:exp}), which is not addressed by previous approaches.

%HERE
\subsection{Context-aware Encoder-decoder}
Motivated by the potential of the Context Encoders \cite{Pathak_CVPR_2016}, 
our CNN learns feature representations of input images via an encoder and reconstruct the harmonized output results through a decoder.
While the proposed deep network bears some resemblance, we add novel components for image harmonization. 
In the following, we present the objective function and  
proposed network architecture with discussion of novel components.

{\flushleft {\bf Objective Function.}}
Given a RGB image $I \in \mathbb{R}^{H \times W \times 3}$ and a provided binary mask $M \in \mathbb{R}^{H \times W \times 1}$ of the composite foreground region, we form the input $X \in \mathbb{R}^{H \times W \times 4}$ by concatenating $I$ and $M$, where $H$ and $W$ are image dimensions.
Our objective is to predict an output image $\hat{Y} = \mathcal{F}(X)$ that optimizes the reconstruction ($L2$) loss with respect to the ground truth image $Y$:
\begin{equation}
\mathcal{L}_{rec}(X) = \frac{1}{2}\sum_{h,w}\parallel Y_{h,w} - \hat{Y}_{h,w} \parallel_2^2.
\label{eq:l2}
\end{equation}
Since the $L2$ loss is optimized with the mean of the data distribution, the results 
are often blurry and thus miss important details and textures from the input image.
To overcome these problems, we show that adding skip links 
from the encoder to the decoder can recover those image details in the proposed network. 

{\flushleft {\bf Network Architecture.}}
Figure \ref{fig:network} shows basic components of our network architecture
with an encoder and a harmonization decoder.
The encoder is a series of convolutional layers and a fully connected layer to learn feature representations from low-level image details to high-level context information.
Note that as we do not have any pooling layers, 
fine details are preserved in the encoder \cite{Pathak_CVPR_2016}.
The decoder is a series of deconvolutional layers 
which aim to reconstruct the image via up-sampling from the representations learned in the encoder and simultaneously adjust the appearances of the foreground region.

However, image details and textures may be lost during the compression process in the encoder, and thus there is less information to reconstruct the contents of the input image.
To retain those details, it is crucial that 
we add a skip link from each convolutional layer in the encoder to each corresponding deconvolutional layer in the decoder.
We show this method is effectively useful without adding additional burdens for training the network.
Furthermore, it can alleviate the problem of the $L2$ loss that prefers a blurry image solution. 

{\flushleft {\bf Implementation Details.}}
We implement the proposed network in Caffe \cite{jia2014caffe} and use the stochastic gradient descent solver for optimization with a fixed learning rate $10^{-8}$.
In addition, we compute the loss on the entire image rather than the foreground mask to account for the reconstruction differences in the background region.
We also try a weighted loss that considers the foreground region more importantly, but
 the results are similar and thus we use a simple loss function. 
 %
%Thus, for simplicity we use a loss functio.
%
Since the entire network is trained from scratch, we use the batch normalization \cite{Ioffe_ICML_2015} followed by a scaling layer and an ELU layer \cite{Clevert_ICLR_2016} after each convolutional and deconvolutional layers to facilitate the training process.

{\flushleft {\bf Discussions.}}
We conduct experiments using the proposed network architecture with different input sizes. 
Interestingly, we find that the one with larger input size 
performs better in practice, and thus we use input resolution of $512 \times 512$.
This observation also matches our intuition when designing the encoder-decoder architecture with skip links, where the network can learn more context information and details from a larger input image.
To generate higher resolution results, we can up-sample the output of the network with joint bilateral filtering \cite{Petschnigg_SIGGRAPH_2004}, in which the input composite image is used as the guidance to keep clear details and sharp textures.
\begin{figure*}[t]
	\centering
	\includegraphics[width=1.0\linewidth]{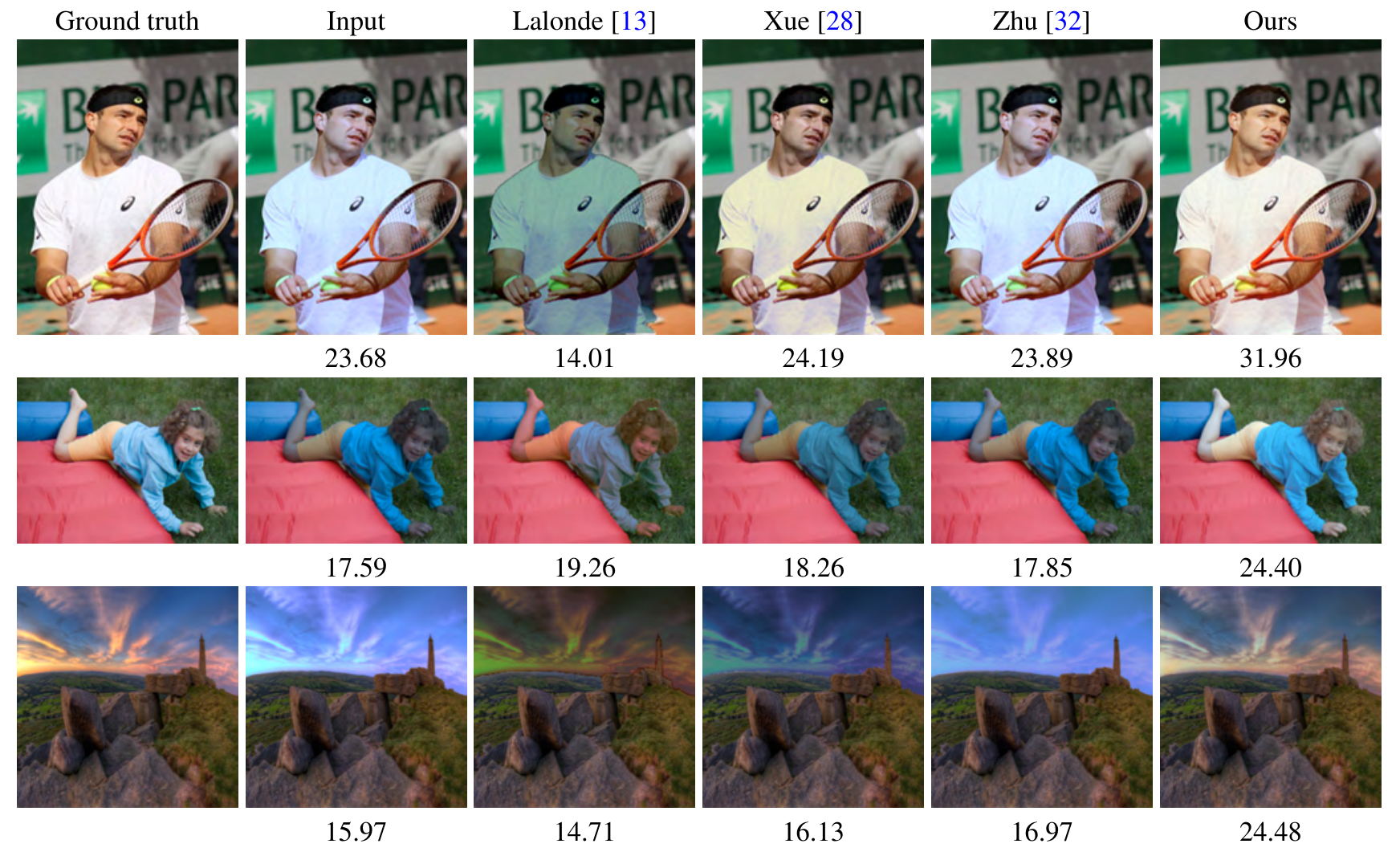}
%	\vspace{2mm}
	\caption{Example results on synthesized datasets for the input, ground truth, three state-of-the-art methods and our proposed network. From the first row to the third one, we show one example for the MSCOCO, MIT-Adobe and Flickr datasets. Each result is associated with a PSNR score. Among all the methods, our harmonization results obtain the highest score.
	}
	\label{fig:synthesize}
%	\vspace{5mm}
\end{figure*}
\subsection{Joint Training with Semantics}
In the previous section, we propose an encoder-decoder network architecture for image harmonization.
In order to further improve harmonization results, it is natural to consider semantics of the composite foreground region. 
The ensuing question is how to incorporate such semantics in our CNN, so that the entire network is still end-to-end trainable.
In this section, we propose a modified network that can jointly train the image harmonization and scene parsing tasks simultaneously, while propagating semantics to improve harmonization results.
The overall architecture is depicted in Figure \ref{fig:network}, which adds the scene parsing decoder branch.

{\flushleft {\bf Joint Loss.}}
In addition to the reconstruction loss described for image harmonization in \eqref{eq:l2}, we introduce a pixel-wise cross-entropy loss with the standard softmax function $\mathbb{E}$ for scene parsing:
\begin{equation}
\mathcal{L}_{cro}(X) = -\sum_{h,w} \log (\mathbb{E}(X_{h,w};\theta)).
\label{eq:softmax}
\end{equation}
We then define a combined loss for both tasks and optimize it jointly:
\begin{equation}
\mathcal{L} = \lambda_{1} \mathcal{L}_{rec} + \lambda_{2}\mathcal{L}_{cro},
\label{eq:combine}
\end{equation}
where $\lambda_{i}$ is the weight to control the balance between losses for image harmonization and scene parsing.

{\flushleft {\bf Network Architecture.}}
We design the joint network by inheriting the encoder-decoder architecture described in the previous section.
Specifically, we add a decoder to predict scene parsing results, while the encoder is to learn feature representations and is shared for both decoders.
To extract semantic knowledge from the scene parsing model and help harmonization process, we concatenate feature maps from each deconvolutional layer of the scene parsing decoder to the harmonization decoder, except for the last layer which focuses on image reconstruction.
In addition, skips links \cite{Long_CVPR_2015} are also connected to the scene parsing decoder to gain more information from the encoder.
{\flushleft {\bf Implementation Details.}}
To enable the training process for the proposed joint network, both the ground truth images for harmonization and scene parsing are required.
We then use a subset of the ADE20K dataset \cite{Zhou_corr_2016}, which contains $12080$ training images with the top $25$ frequent labels.
Similarly, training pairs for harmonization are obtained in a way described in the data acquisition section via color transfer.

To train the joint network, we start with the training data from the ADE20K dataset to obtain an initial solution for both the harmonization and scene parsing by optimizing \eqref{eq:combine}.
We set $\lambda_1 = 1$ and $\lambda_2 = 100$ with a fixed learning rate $10^{-8}$.
Next, we  fix the scene parsing decoder with $\lambda_2 = 0$ and finetune rest of the network using all the training data introduced in Section \ref{sec:data} to achieve the optimal solution for image harmonization.
Note that, during this fintuning step, the scene parsing decoder is able to propagate learned semantic information through the links between two decoders.
{\flushleft {\bf Discussions.}}
With the incorporated scene parsing model, our network can learn the color distribution of certain semantic categories, e.g., the skin color on human or the sky-like colors.
In addition, the learned background semantics can help identify which region to match for better foreground adjustment.
During harmonization, it essentially uses these learned semantic priors to improve the realism of output results.
Moreover, the incorporation of semantic information through joint training not only helps our image harmonization task, but also can be adopted to benefit other image editing tasks \cite{Zhang_ECCV_2016, Pathak_CVPR_2016}.
\begin{table}[t]
	\caption{Comparisons of methods with mean-squared errors (MSE) on three synthesized datasets.
	}
	\vspace{1mm}
	\small
	\centering
	\renewcommand{\arraystretch}{1.5}
	\setlength{\tabcolsep}{6pt}     
	\begin{tabular}{|c|c|c|c|}
		\hline
		& MSCOCO & MIT-Adobe & Flickr \\
		\hline
		\vspace{-2mm}
		
		cut-and-paste & 400.5 & 552.5 & 701.6 \\
		\vspace{-2mm}
		
		Lalonde \cite{Lalonde_ICCV_2007} & 667.0 & 1207.8 & 2371.0 \\
		\vspace{-2mm}
		
		Xue \cite{Xue_siggraph_2012} & 351.6 & 568.3 & 785.1 \\
		\vspace{-2mm}
		
		Zhu \cite{Zhu_ICCV_2015} & 322.2 & 360.3 & 475.9 \\
		\vspace{-2mm}
		
		Ours (w/o semantics) & 80.5 & 168.8 & 491.7 \\
		
		Ours & \textbf{76.1} & \textbf{142.8} & \textbf{406.8} \\
		
		\hline
	\end{tabular}
	\label{tab:mse}
	\vspace{-2mm}
\end{table}

To validate our scene parsing model, we compare the proposed joint network to a deeplab model \cite{Chen_ICLR_2015}, MSc-COCO-LargeFOV, that has a similar model capacity and size to our model but is initialized from a pre-trained model for semantic segmentation.
We evaluate the scene parsing results on the validation set of the ADE20K dataset with the top $25$ frequent labels. 
The mean intersection-over-union (IoU) accuracy of our joint network is $32.2$, while the MSc-COCO-LargeFOV model achieves IoU as $36.0$.
Although our model is not specifically designed for scene parsing and is learned from scratch, it shows that our method performs competitively against a state-of-the-art model for semantic segmentation.
\begin{table}[t]
	\caption{Comparisons of methods with PSNR scores on three synthesized datasets.
	}
	\vspace{1mm}
	\small
	\centering
	\renewcommand{\arraystretch}{1.5}
	\setlength{\tabcolsep}{6pt}     
	\begin{tabular}{|c|c|c|c|}
		\hline
		& MSCOCO & MIT-Adobe & Flickr \\
		\hline
		\vspace{-2mm}
		
		cut-and-paste & 26.3 & 23.9 & 25.9 \\
		\vspace{-2mm}
		
		Lalonde \cite{Lalonde_ICCV_2007} & 22.7 & 21.1 & 18.9 \\
		\vspace{-2mm}
		
		Xue \cite{Xue_siggraph_2012} & 26.9 & 24.6 & 25.0 \\
		\vspace{-2mm}
		
		Zhu \cite{Zhu_ICCV_2015} & 26.9 & 25.8 & 25.4 \\
		\vspace{-2mm}
		
		Ours (w/o semantics) & 32.2 & 27.5 & 27.2 \\
		
		Ours & \textbf{32.9} & \textbf{28.7} & \textbf{27.4} \\
		
		\hline
	\end{tabular}
	\label{tab:psnr}
%	\vspace{-2mm}
\end{table}
\begin{figure}[t]
	\centering
	\begin{tabular}
		{@{\hspace{0mm}}c@{\hspace{1mm}} @{\hspace{0mm}}c@{\hspace{1mm}} @{\hspace{0mm}}c@{\hspace{0mm}}
		}
		
		Input & No semantics & With semantics \\
		
		\includegraphics[width=0.32\linewidth]{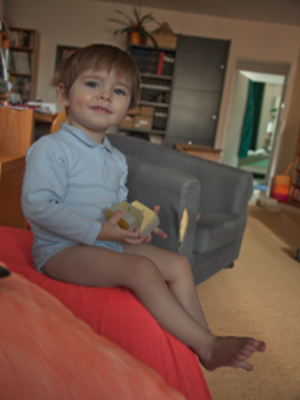} &
		\includegraphics[width=0.32\linewidth]{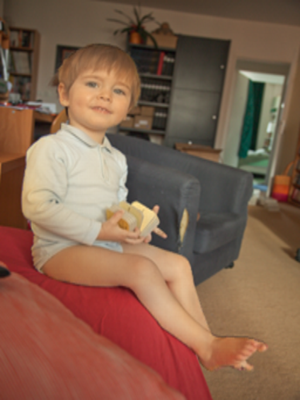} &
		\includegraphics[width=0.32\linewidth]{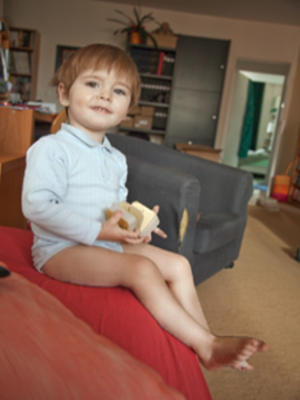} \\
		
		18.86 & 28.15 & 33.32 \\
		
	\end{tabular}
	\vspace{2mm}
	\caption{Example results to show the comparison of our network with or without incorporating semantic information.
		With semantics, our result can recover the skin color and obtain higher PSNR score.
	}
	\label{fig:semantic}
	\vspace{-2mm}
\end{figure}
\section{Experimental Results}
\label{sec:exp}

We present the main results on image harmonization with comparisons to the state-of-the-art methods in this section.
More results and analysis can be found in the supplementary material.

{\flushleft {\bf Synthesized Data.}}
We first evaluate the proposed method on our synthesized dataset for quantitative comparisons.
Table \ref{tab:mse} and \ref{tab:psnr} show the results of mean-squared errors (MSE) and PSNR scores between the ground truth and harmonized image.
Note that it is the first quantitative evaluation on image harmonization, which reflects how close different results are to realistic images. 
We show that our joint network consistently achieves better performance compared to the single network without combining scene parsing decoder and other state-of-the-art algorithms \cite{Lalonde_ICCV_2007, Xue_siggraph_2012, Zhu_ICCV_2015} on all three synthesized datasets in terms of MSE and PSNR.
In addition, it is also worth noticing that our baseline network without semantics already outperforms other existing methods.
%

%To validate the effectiveness of our joint training scheme, we try to separately incorporate a state-of-the-art scene parsing model into our single encoder-decoder harmonization framework to provide semantic information.
%%
%Surprisingly, the performance is worse than our joint model, where our scene parsing decoder is learned from scratch.
%%
%More details and analysis are provided in the supplementary material.

In Figure \ref{fig:synthesize}, we show visual comparisons with respect to PSNR of the harmonization results generated from different methods.
Overall, the harmonized images by the proposed methods are more realistic and closer to the ground truth images, with higher PSNR values. 
In addition, Figure \ref{fig:semantic} presents one comparison of our networks with and without incorporating the scene parsing decoder.
With semantic understandings, our joint network is able to harmonize foreground regions according to their semantics and produce realistic appearance adjustments, while the one without semantics may generate unsatisfactory results in some cases.
{\flushleft {\bf Real Composite Images.}}
To evaluate the effectiveness of the proposed joint network in real scenarios, we create a test set of $52$ real composite images and combine $48$ examples from 
Xue et al. \cite{Xue_siggraph_2012}, resulting in a total of $100$ high-quality composite images.
To cover a variety of real examples, we create composite images including various scenes and stylized images, where the composite foreground region can be an object or a scene.

We follow the same procedure as \cite{Xue_siggraph_2012, Zhu_ICCV_2015} to set up a user study on Amazon Mechanical Turk, in which each user sees two randomly selected results at a time and is asked to choose the one that looks more realistic.
For sanity checks, we use ground truth images from the synthesized dataset and heavily edited images to create easily distinguishable pairs that are used to filter out bad users.
As a result, a total of $225$ subjects participate in this study with a total of $10773$ pairwise results ($10.8$ results for each pair of different methods on average).
After obtaining all the pairwise results, we use the Bradley-Terry model (B-T model) \cite{BradleyTerry, Lai_CVPR_2016} to calculate the global ranking score for each method.
\begin{figure}[t]
	\centering
	\begin{tabular}
		{@{\hspace{0mm}}c@{\hspace{1mm}} @{\hspace{0mm}}c@{\hspace{1mm}} @{\hspace{0mm}}c@{\hspace{0mm}}
		}
		\includegraphics[width=0.32\linewidth]{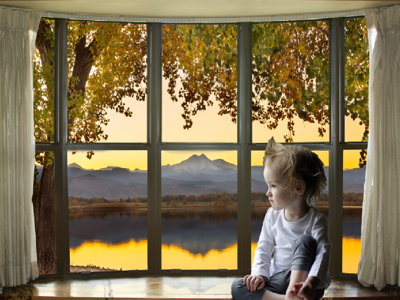} &
		\includegraphics[width=0.32\linewidth]{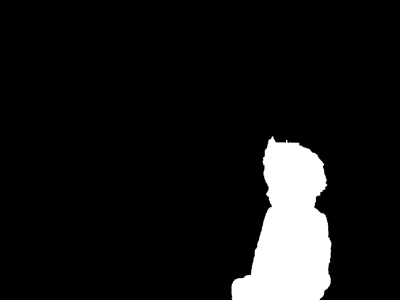} &
		\includegraphics[width=0.32\linewidth]{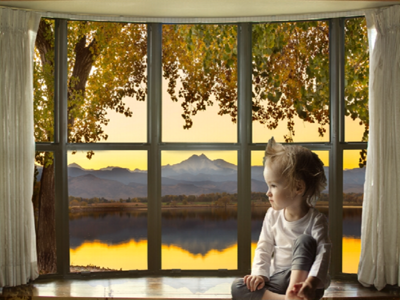} \\
		%		\vspace{1mm}
		(a) Input & & \\
		&
		\includegraphics[width=0.32\linewidth]{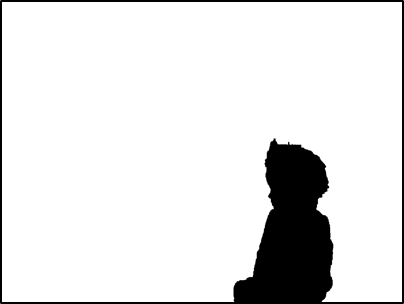} &
		\includegraphics[width=0.32\linewidth]{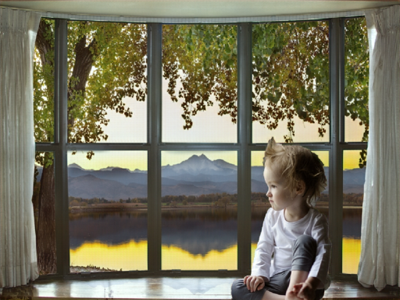} \\
		& (b) Mask & (c) Output \\
		%		\vspace{1mm}
	\end{tabular}
	\vspace{2mm}
	\caption{Given an input image (a), our network can adjust the foreground region according to the provided mask (b) and produce the output (c).
		In this example, we invert the mask from the one in the first row to the one in the second row, and generate harmonization results that account for different context and semantic information.
	}
	\label{fig:guidance}
	\vspace{1mm}
\end{figure}
\begin{table}[t]
	\caption{Comparisons of methods with B-T scores on real composite datasets.
	}
	\vspace{1mm}
	\small
	\centering
	\renewcommand{\arraystretch}{1.5}
	\setlength{\tabcolsep}{10pt}     
	\begin{tabular}{|c|c|c|c|}
		\hline
		Dataset & \cite{Xue_siggraph_2012} & Our test set & Overall \\
		\hline
		\hline
		\vspace{-2mm}
		
		cut-and-paste & 1.080 & 1.168 & 1.139 \\
		\vspace{-2mm}
		
		Lalonde \cite{Lalonde_ICCV_2007} & 0.557 & 0.067 & 0.297 \\
		\vspace{-2mm}
		
		Xue \cite{Xue_siggraph_2012} & 1.130 & 0.885 & 1.002 \\
		\vspace{-2mm}
		
		Zhu \cite{Zhu_ICCV_2015} & 0.875 & 0.867 & 0.876 \\
		
		Ours & \textbf{1.237} & \textbf{1.568} &  \textbf{1.424} \\
		
		\hline
	\end{tabular}
	\label{tab:real}
	\vspace{-3mm}
\end{table}
\begin{figure*}[t]
	\centering
	\includegraphics[width=1.0\linewidth]{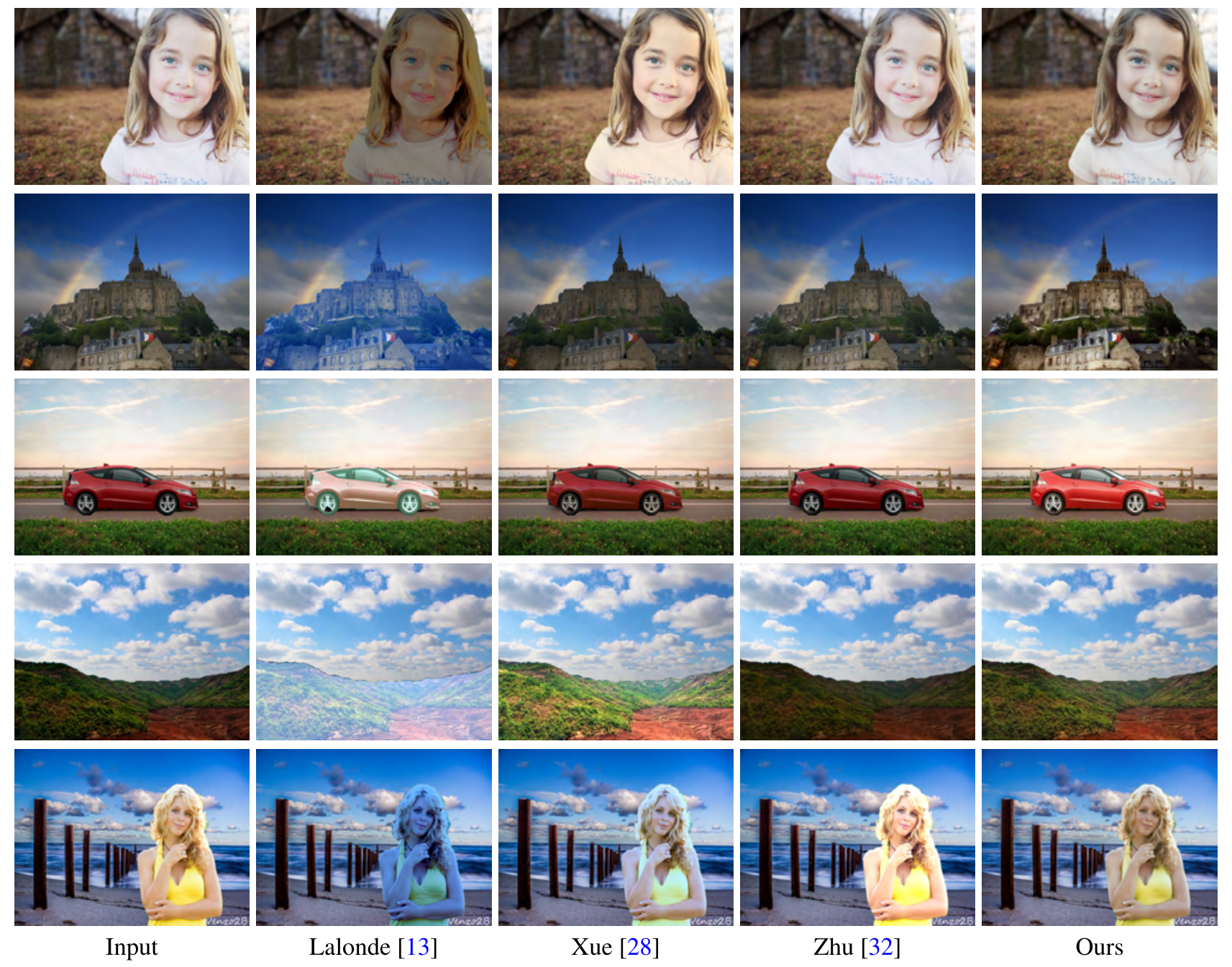}
%	\vspace{2mm}
	\caption{Example results on real composite images for the input, three state-of-the-art methods and our proposed network.
		We show that our method produces realistic harmonized images by adjusting composite foreground regions containing various scenes or objects.
	}
	\label{fig:real}
	\vspace{2mm}
\end{figure*}

Table \ref{tab:real} shows that our method achieves the highest B-T score in terms of realism compared to state-of-the-art approaches on both our created test set and examples from \cite{Xue_siggraph_2012}.
Interestingly, our method is the only one that can improve the harmonization result with a significant margin from the input image (by cut-and-paste).
Figure \ref{fig:real} shows sample harmonized images by the evaluated methods.
Overall, our joint network produces realistic output images, which validates the effectiveness of using synthesized data to directly learn how to harmonize composite images from realistic ground truth images. 
The results from \cite{Xue_siggraph_2012} may be easily affected by the large appearance difference between the background and foreground regions during matching.
For the method \cite{Zhu_ICCV_2015}, it may generate unsatisfactory results due to the errors introduced during realism prediction, which may affect the color optimization step.
In contrast, our network adopts a single feed-forward scheme learned from a well-constructed training set, and utilizes semantic information to improve harmonization results.
The complete results on the real composite test set are presented in the supplementary material.
%
%\section{Network Properties}
%
{\flushleft {\bf Generalization to Background Masks.}}
With the provided foreground mask, our network can learn context and semantic information while transforming the composite image to a realistic output image.
Therefore, our method can be applied to any foreground masks containing arbitrary objects, scenes or clutter backgrounds.
Figure \ref{fig:guidance} illustrates one example, where originally the adjusted foreground region is the \textit{child}.
Instead, we can invert the mask and focus on harmonizing the region of \textit{inverted child}.
The result shows that our network can produce realistic outputs from different foreground masks.
%
%{\flushleft {\bf Iterative Harmonization.}}
%%
%Another interesting property of our network is that we can iteratively apply the network on a composite image until convergence or determine how much degree we would like to harmonize.
%%
%For instance, given an input image, we can output the harmonized result and use it as a new input to run the harmonization again (see Figure \ref{fig:iterative} as an example).
%%
%In this way, our network gradually approaches to a solution that the foreground distribution matches the background contexts.
%%
%Although there is no theoretical proof, we do observe that the solution usually converges in a reasonable way.
%%
%In addition, from users' perspective, it allows them to easily select the one they prefer among a set of harmonized results.
%%
%%
%\begin{figure}[t]
%	\centering
%	\begin{tabular}
%		{@{\hspace{0mm}}c@{\hspace{1mm}} @{\hspace{0mm}}c@{\hspace{1mm}} @{\hspace{0mm}}c@{\hspace{0mm}}
%		}
%		\includegraphics[width=0.32\linewidth]{figure/iterative/input.png} &
%		\includegraphics[width=0.32\linewidth]{figure/iterative/output_iter1.png} &
%		\includegraphics[width=0.32\linewidth]{figure/iterative/output_iter3.png} \\
%		\vspace{1mm}
%		(a) Input & (b) Iteration 1 & (c) Iteration 3 \\
%	\end{tabular}
%	\caption{Given an input image (a), we can iteratively apply our network to produce results with different degrees of harmonization effects.
%	%
%	For instance, the harmonized \textit{dog} has a whiter look that closely matches the background context.
%	}
%	\label{fig:iterative}
%\end{figure}
%
%
{\flushleft {\bf Runtime Performance.}}
Previous image harmonization methods rely on matching statistics \cite{Lalonde_ICCV_2007, Xue_siggraph_2012} or optimizing an adjustment function \cite{Zhu_ICCV_2015}, which usually require longer processing time (more than $10$ seconds with a 3.4GHz Core Xeon CPU) on a $512 \times 512$ test image.
In contrast, our proposed CNN is able to harmonize an image in 0.1 seconds with a Titan X GPU and
12GB memory, or $3$ seconds with a CPU.
%
%This enables the applicability of applying our network on harmonizing videos.

\section{Concluding Remarks}
In this paper, we present a novel network that can capture both the context and semantic information for image harmonization.
We demonstrate that our joint network can be trained in an end-to-end manner, where the semantic decoder branch can effectively provide semantics to help harmonization.
In addition, to facilitate the training process, we develop an efficient method to collect large-scale and high-quality training pairs.
Experimental results show that our method performs favorably on both the synthesized datasets and real composite images against other state-of-the-art algorithms.

\clearpage
{\small
\bibliographystyle{ieee}
\bibliography{mybib}

\begin{thebibliography}{10}\itemsep=-1pt

\bibitem{Bengio_PAMI_2013}
Y.~Bengio, A.~Courville, and P.~Vincent.
\newblock Representation learning: A review and new perspectives.
\newblock {\em PAMI}, 35(8):1798 -- 1828, 2013.

\bibitem{BradleyTerry}
R.~A. Bradley and M.~E. Terry.
\newblock Rank analysis of incomplete block designs: I. the method of paired
  comparisons.
\newblock {\em Biometrika}, 39:324--345, 1952.

\bibitem{Bychkovsky_CVPR_2011}
V.~Bychkovsky, S.~Paris, E.~Chan, and F.~Durand.
\newblock Learning photographic global tonal adjustment with a database of
  input/output image pairs.
\newblock In {\em CVPR}, 2011.

\bibitem{Chen_ICLR_2015}
L.-C. Chen, G.~Papandreou, I.~Kokkinos, K.~Murphy, and A.~L. Yuille.
\newblock Semantic image segmentation with deep convolutional nets and fully
  connected crfs.
\newblock In {\em ICLR}, 2015.

\bibitem{Clevert_ICLR_2016}
D.~Clevert, T.~Unterthiner, and S.~Hochreiter.
\newblock Fast and accurate deep network learning by exponential linear units
  (elus).
\newblock In {\em ICLR}, 2016.

\bibitem{Hwang_ECCV_2012}
S.~J. Hwang, A.~Kapoor, and S.~B. Kang.
\newblock Context-based automatic local image enhancement.
\newblock In {\em ECCV}, 2012.

\bibitem{Iizuka_SIGGRAPH_2016}
S.~Iizuka, E.~Simo-Serra, and H.~Ishikawa.
\newblock Let there be color!: Joint end-to-end learning of global and local
  image priors for automatic image colorization with simultaneous
  classification.
\newblock {\em ACM Trans. Graph. (proc. SIGGRAPH)}, 35(4), 2016.

\bibitem{Ioffe_ICML_2015}
S.~Ioffe and C.~Szegedy.
\newblock Batch normalization: Accelerating deep network training by reducing
  internal covariate shift.
\newblock In {\em ICML}, 2015.

\bibitem{jia2014caffe}
Y.~Jia, E.~Shelhamer, J.~Donahue, S.~Karayev, J.~Long, R.~Girshick,
  S.~Guadarrama, and T.~Darrell.
\newblock Caffe: Convolutional architecture for fast feature embedding.
\newblock {\em arXiv preprint arXiv:1408.5093}, 2014.

\bibitem{Johnson_TVGG_2011}
M.~K. Johnson, K.~Dale, S.~Avidan, H.~Pfister, W.~T. Freeman, and W.~Matusik.
\newblock Cg2real: Improving the realism of computer generated images using a
  large collection of photographs.
\newblock {\em IEEE Trans. Vis. Comp. Graph.}, 17(9), 2011.

\bibitem{Kong_ECCV_2016}
S.~Kong, X.~Shen, Z.~Lin, R.~Mech, and C.~Fowlkes.
\newblock Photo aesthetics ranking network with attributes and content
  adaptation.
\newblock In {\em ECCV}, 2016.

\bibitem{Lai_CVPR_2016}
W.-S. Lai, J.-B. Huang, Z.~Hu, N.~Ahuja, and M.-H. Yang.
\newblock A comparative study for single image blind deblurring.
\newblock In {\em CVPR}, 2016.

\bibitem{Lalonde_ICCV_2007}
J.-F. Lalonde and A.~A. Efros.
\newblock Using color compatibility for assessing image realism.
\newblock In {\em ICCV}, 2007.

\bibitem{Larsson_ECCV_2016}
G.~Larsson, M.~Maire, and G.~Shakhnarovich.
\newblock Learning representations for automatic colorization.
\newblock In {\em ECCV}, 2016.

\bibitem{Lazebnik_CVPR_2006}
S.~Lazebnik, C.~Schmid, and J.~Ponce.
\newblock Beyond bags of features: Spatial pyramid matching for recognizing
  natural scene categories.
\newblock In {\em CVPR}, 2006.

\bibitem{Lee_CVPR_2016}
J.-Y. Lee, K.~Sunkavalli, Z.~Lin, X.~Shens, and I.~S. Kweon.
\newblock Automatic content-aware color and tone stylization.
\newblock In {\em CVPR}, 2016.

\bibitem{Lin_ECCV_2014}
T.-Y. Lin, M.~Maire, S.~Belongie, J.~Hays, P.~Perona, D.~Ramanan,
  P.~Doll{\'a}r, and C.~L. Zitnick.
\newblock Microsoft {COCO}: Common objects in context.
\newblock In {\em ECCV}, 2014.

\bibitem{Liu_ECCV_2016}
S.~Liu, J.~Pan, and M.-H. Yang.
\newblock Learning recursive filters for low-level vision via a hybrid neural
  network.
\newblock In {\em ECCV}, 2016.

\bibitem{Long_CVPR_2015}
J.~Long, E.~Shelhamer, and T.~Darrell.
\newblock Fully convolutional networks for semantic segmentation.
\newblock In {\em CVPR}, 2015.

\bibitem{Pathak_CVPR_2016}
D.~Pathak, P.~Kr{\"{a}}henb{\"{u}}hl, J.~Donahue, T.~Darrell, and A.~A. Efros.
\newblock Context encoders : Feature learning by inpainting.
\newblock In {\em CVPR}, 2016.

\bibitem{Perez_siggraph_2003}
P.~P{\'{e}}rez, M.~Gangnet, and A.~Blake.
\newblock Poisson image editing.
\newblock {\em ACM Trans. Graph. (proc. SIGGRAPH)}, 22(3), 2003.

\bibitem{Petschnigg_SIGGRAPH_2004}
G.~Petschnigg, R.~Szeliski, M.~Agrawala, M.~Cohen, H.~Hoppe, and K.~Toyama.
\newblock Digital photography with flash and no-flash image pairs.
\newblock {\em ACM Trans. Graph. (proc. SIGGRAPH)}, 23(3), 2004.

\bibitem{Pitie_CVMP_2007}
F.~Piti\'{e} and A.~Kokaram.
\newblock The linear monge-kantorovitch linear colour mapping for example-based
  colour transfer.
\newblock In {\em CVMP}, 2007.

\bibitem{Reinhard_CGA_2001}
E.~Reinhard, M.~Ashikhmin, B.~Gooch, and P.~Shirley.
\newblock Color transfer between images.
\newblock {\em IEEE Comp. Graph. Appl.}, 21(5):34--41, 2001.

\bibitem{Sunkavalli_siggraph_2010}
K.~Sunkavalli, M.~K. Johnson, W.~Matusik, and H.~Pfister.
\newblock Multi-scale image harmonization.
\newblock {\em ACM Trans. Graph. (proc. SIGGRAPH)}, 29(4), 2010.

\bibitem{Tao_IJCV_2013}
M.~W. Tao, M.~K. Johnson, and S.~Paris.
\newblock Error-tolerant image compositing.
\newblock {\em IJCV}, 103(2):178--189, 2013.

\bibitem{Tsai_SIGGRAPH_2016}
Y.-H. Tsai, X.~Shen, Z.~Lin, K.~Sunkavalli, and M.-H. Yang.
\newblock Sky is not the limit: Semantic-aware sky replacement.
\newblock {\em ACM Trans. Graph. (proc. SIGGRAPH)}, 35(4), 2016.

\bibitem{Xue_siggraph_2012}
S.~Xue, A.~Agarwala, J.~Dorsey, and H.~Rushmeier.
\newblock Understanding and improving the realism of image composites.
\newblock {\em ACM Trans. Graph. (proc. SIGGRAPH)}, 31(4), 2012.

\bibitem{Yan_siggraph_2015}
Z.~Yan, H.~Zhang, B.~Wang, S.~Paris, and Y.~Yu.
\newblock Automatic photo adjustment using deep neural networks.
\newblock {\em ACM Trans. Graph.}, 2015.

\bibitem{Zhang_ECCV_2016}
R.~Zhang, P.~Isola, and A.~A. Efros.
\newblock Colorful image colorization.
\newblock In {\em ECCV}, 2016.

\bibitem{Zhou_corr_2016}
B.~Zhou, H.~Zhao, X.~Puig, S.~Fidler, A.~Barriuso, and A.~Torralba.
\newblock Semantic understanding of scenes through the {ADE20K} dataset.
\newblock {\em CoRR}, abs/1608.05442, 2016.

\bibitem{Zhu_ICCV_2015}
J.-Y. Zhu, P.~Kr{\"{a}}henb{\"{u}}hl, E.~Shechtman, and A.~A. Efros.
\newblock Learning a discriminative model for the perception of realism in
  composite images.
\newblock In {\em ICCV}, 2015.

\end{thebibliography}
}

\end{document}